\def\year{2022}\relax
\documentclass[letterpaper]{article} 
\usepackage{aaai22}  
\usepackage{times}  
\usepackage{helvet}  
\usepackage{courier}  
\usepackage[hyphens]{url}  
\usepackage{graphicx} 
\usepackage{natbib}  
\usepackage{caption} 
\DeclareCaptionStyle{ruled}{labelfont=normalfont,labelsep=colon,strut=off} 
\usepackage{algorithm}
\usepackage{algorithmic}

\usepackage{newfloat}
\usepackage{listings}
\floatstyle{ruled}
\newfloat{listing}{tb}{lst}{}
\floatname{listing}{Listing}
\usepackage{color}
\usepackage{xcolor}
\newcommand{\edit}[1]{{#1}}

\title{Qualitative Analysis of a Graph Transformer Approach to Addressing Hate Speech:
Adapting to Dynamically Changing Content}
\author{
    Liam Hebert,
    Hong Yi Chen,
    Robin Cohen, 
    Lukasz Golab
}
\affiliations{
    University of Waterloo \\
    \{liam.hebert, hong.yi.chen, rcohen, lgolab\}@uwaterloo.ca

}

\begin{document}

\maketitle

\begin{abstract}
Our work advances an approach for predicting hate speech in social media, drawing out the critical need to consider the discussions that follow a post to successfully detect when hateful discourse may arise. Using graph transformer networks, coupled with modelling attention and BERT-level natural language processing, our approach can capture context and anticipate upcoming anti-social behaviour. In this paper, we offer a detailed qualitative analysis of this solution for hate speech detection in social networks, leading to insights into where the method has the most impressive outcomes in comparison with competitors and identifying scenarios where there are challenges to achieving ideal performance. Included is an exploration of the kinds of posts that permeate social media today, including the use of hateful images. This suggests avenues for extending our model to be more comprehensive. A key insight is that the focus on reasoning about the concept of context positions us well to be able to support multi-modal analysis of online posts. We conclude with a reflection on how the problem we are addressing relates especially well to the theme of dynamic change, a critical concern for all AI solutions for social impact. We also comment briefly on how mental health well-being can be advanced with our work, through curated content attuned to the extent of hate in posts.
\end{abstract}

\section{Introduction}
Online social platforms have allowed vast amounts of communication between individuals at an unprecedented scale. Platforms such as Facebook have over 2.9 billion monthly active users who share opinions and connect with other users\footnote{https://www.statista.com/statistics/264810/number-of-monthly-active-facebook-users-worldwide/}. A central tenet of these platforms is the removal of traditional editorial barriers to reach a wider audience. Opinions or commentary do not need to be regulated by editors before they can be published and shared. However, this open approach to free speech has also led to the explosion of propaganda, violence, and abuse against users based on their race, gender, and religion \cite{das2020hate}. In addition, widespread dissemination of hateful speech has resulted in traumatizing mental health effects for the victims \cite{janikke2019hate} and has ignited social tensions and polarization between groups \cite{Waller2021}. To combat this trend, social platforms have created rigorous community guidelines which describe the kinds of content that can be shared\footnote{https://transparency.fb.com/policies/community-standards/}. These guidelines are then enforced by teams of human moderators who manually allow or disallow content. While effective, this approach can be insufficient for coping with the growing scale of these platforms. 

In an effort to allow improvements, platforms have also turned to the use of automated methods to detect hate speech \cite{das2022data, mathew2021hatexplain}, aiming to classify the text that comprises the comment as either hate speech or non-hate speech. However, we argue that this comment-only scope is becoming increasingly limited and ineffective, due to the importance of capturing context when deciding whether speech is hateful or not.

To this end, we have designed an approach that goes beyond current hate speech labelling efforts in three distinct ways \cite{hebert2022predicting}. First, we analyze entire discussions following a post, to detect hate speech. Second, we support predicting when hate speech will occur, rather than simply reacting to hateful posts once they are detected. 
All of this is achieved using graph transformer networks, coupled with modelling attention and BERT natural language processing. In so doing, we can capture the discussion context and anticipate upcoming anti-social behaviour. 
This allows us to analyze the conversational dynamics of different communities, being sensitive to cases where the usage of a slur can be re-appropriated to appear to be non-abusive. For example, the usage of certain slurs has been largely re-appropriated in African American culture as a normal part of their vernacular \cite{thomas2007phonological}. 

In this paper, we offer a detailed qualitative analysis of this solution for hate speech detection in social networks, leading to insights into where the method has the most impressive outcomes in comparison with competitors and identifying scenarios where there are challenges to achieving ideal performance. We draw out the key observation that comments on social platforms have evolved to include images and external articles. These additional elements can provide essential context to properly understanding the content that follows. 

We will conclude with a discussion on how to extend \edit{our method} to encompass the processing of images, towards the holistic processing of online posts. We will also return to the concern of mental health well-being and discuss how our more comprehensive automated solution for hate speech prediction can be the basis for some significant steps forward. The social impact that we anticipate coming from the research presented in this paper will be on social media environments and their users.

\section{Data and Methods}

\subsection{Comment-Only Hate Speech Models}
To evaluate recent work in hate speech detection, we selected MuRIL by \citet{das2022data} and Bert-HateXplain by \citet{mathew2021hatexplain}. Both of these systems are based on the BERT transformer architecture, which can create rich embeddings of text toward classification tasks \cite{devlin2019bert}. We refer to these methods as comment-only hate speech models. 

The main difference between the two methods is the data that both systems were trained on. For Bert-HateXplain, the authors collected a combined dataset of 20,148 hateful tweets and posts from social platforms Twitter and Gab. For MuRIL, the authors combined the HateXplain dataset with \citet{founta2018large} (85,775 tweets) and \citet{davidson2017automated} (24,783 tweets). This combined approach was found to outperform HateXplain to become state-of-the-art in hate speech detection \cite{das2022data}. 

\subsection{Graph Hate Speech Models}
To study the usage of Graph Networks for hate speech detection, we focus on our Graphormer approach proposed in \citet{hebert2022predicting}. This model was novel in its ability to advance the study of hate speech detection in social media by a) predicting where hate may arise rather than simply reacting to posts that have been labelled as hate and b) leveraging graph transformers for capturing contextual attention between comments in discussion graphs. 

\begin{figure}
    \centering
    \includegraphics[width=\linewidth]{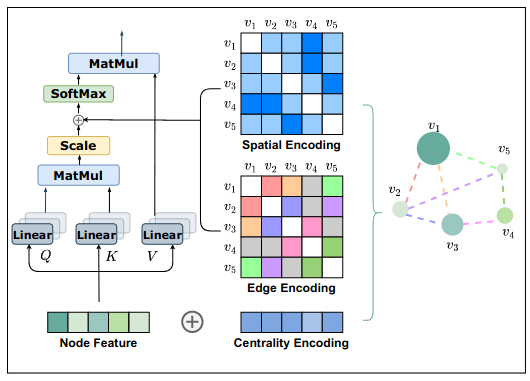}
    \caption{Graphormer Architecture}
    \label{fig:graphormer}
\end{figure}
Core to this model is the Graphormer architecture (Figure \ref{fig:graphormer}), which was originally created by \citet{ying2021transformers} to predict molecular properties. Graphormer uses a transformer model to create embeddings of atoms by computing self-attention relationships between each atom of the molecule in relation to their structure. The key to this approach is that self-attention can be computed between all nodes of a graph irrespective of their structural distance. This is contrasted by previous graph neural networks, which are constrained to computing relationships between immediate neighbours of a node \cite{wu2020comprehensive}. 

To adapt Graphormer to social media analysis, we proposed reformulating hate speech detection as a graph prediction task. Under this approach, comments are represented as nodes and edges are the reply-to relationships between them. We first initiate this graph by creating BERT embeddings of each comment. Then, we aggregate and process the embeddings in relation to the discussion structure using Graphormer, creating hate predictions for each node in the graph. To focus on proactive predictions, the label of each node is an ordinal value (0-4) based on how prevalent and encouraged the hate speech that follows that comment. This training objective requires the model to reason about the degree of hate throughout the entire discussion rather than being isolated to the comment itself.  

In this study, we also evaluate a baseline approach that uses Graph Attention Networks (GAT) \cite{velickovic2018graph}. GAT models have previously been adapted for hate speech detection by \citet{parmentier2019learning, parmentier2021}. Like Graphormer, GAT models utilize attention to create node embeddings in a graph structure. However, this attention is constrained to direct node neighbours, a limitation overcome by Graphormer by utilizing transformers. This can result in predictions that focus more on the immediate discussion context rather than the larger global context.  Other work has examined graph-based approaches for hate speech detection \cite{mishra-etal-2019-abusive, tian-etal-2022-duck}; we confine our attention here to comparisons with the models already described above.

\subsection{Reddit}
We focus on the social platform Reddit to capture examples for this study. On Reddit, discussions take place in topic-oriented communities called subreddits. Each discussion is organized in tree-like structures where users can create branching threads by replying to any comment in the tree. Prior work analyzing Reddit communities has found that these communities exhibit significant differences in their social makeup and communal behaviours \cite{Waller2021}. For example, communities such as r/conservatives demonstrate a right-leaning bias and community whereas r/politics contains a polarizing left-leaning bias. 

We analyzed 32 discussions from 16 different communities centred around contentious topics and unique communities. Each discussion was chosen based on the amount of comments it had. We also draw from Reddit conversations sampled in \citet{kurrek2020towards} for examples of reclaimed speech. For this paper, we select five interesting examples to display in the section that follows.


\section{Analysis}
In this study, we focus on capturing examples from two categories of conversational hate speech. First, we start with samples of contextual hate speech, harmful comments that directly refer to or respond to the prior discussion. Examples of this kind of hate speech would be a harmful commentary on contentious topics, such as responding negatively to gay rights. We hypothesize that comment-only methods would fail to capture the contextual nuances that underpin the hatefulness of the target text. This can result in false positives or false negatives when comments are judged in isolation. 

The second category of hateful speech we study is inciteful hate speech: comments that at first glance appear to be neutral but are designed to prompt harmful discourse by other users. These examples aim to evaluate the ability of graph hate speech models to proactively predict the direction of conversations toward hatefulness, rather than only detect the hate of individual comments. As a result, this direction also evaluates the ability of graph methods to capture the dynamic nature of social media discussions, where conversations are not static but grow over time as users add replies to the content posted by other users. Examples of this kind of speech would be comments concerning US president Biden in right-leaning political subreddits, which can prompt hateful comments and threats.

For each comment in the discussion, we predict labels between zero and four using comment-only and graph hate speech models. To match the ordinal predictions given by Graphormer, we follow \citet{hebert2022predicting} and map the zero to one prediction given by comment-only models to bins of width 0.20 ([0-0.20], [0.20 - 0.40]. [0.40 - 0.60], [0.60-0.80], [0.80-1]). To capture the ability of graph methods to adapt to evolving conversations, we initialize the discussion graph with the initial post and immediate replies (depth 1). We then iteratively predict the labels of each comment by gradually increasing the depth of the discussion tree provided to the graph models. As such, graph methods are constrained to make predictions about the direction of conversations without seeing future comments. 

\subsection{Contextual Hate Speech}
\begin{table*}
    \centering
        \caption{Conversation on r/gay containing contextual speech leading to false positive predictions}
    \begin{tabular}{c|p{8cm}|c|c|c|c}
        Depth & Text & Graphormer & GAT & Bert-HateXplain & MuRIL \\
        \hline
        \hline
        0 & Anyone else loving Lil Nas X meming on biggots? [image] & 2 & 1 & 3 & 0 \\
        \hline
        1 & What was this in reference too? & 1 & 1 & 0 & 0  \\
        \hline 
        2 & Biggots are mad about his most recent song/music video & 1 & 2 & 1 & 0  \\
        \hline
        3 & Ah ok! I'll have to check it out! & 0 & 0 & 0 & 0 \\
        \hline
        4 & The song is called Montero (call me by your name). & 0 & 0 & 0 & 0 \\
        \hline
        5 & My favorite line is "I wanna f*ck you cuz I envy you", and that hit hard & 0 & 0 & 2 & 4 \\
        \hline
        6 & 'I want that jet leg from f*ckin' and flyin' call me in & 0 & 1 & 3 & 4 \\
    \end{tabular}
    \label{tab:contextual-gay}
\end{table*}
We start by analyzing a conversation that took place on the subreddit /r/gay, a community centred around LGBTQ topics (Table \ref{tab:contextual-gay}). In this conversation, users are discussing a tweet from openly gay pop artist Lil Nas X. Our analysis of this discussion thread demonstrates how a lack of context can lead to false predictions about hatefulness. 

The conversation begins with the initial post "Anyone else loving Lil Nas X meming on biggots", referring to his tweet stating "i thought y'all didn't like political correctness. what happened?". This comment resulted in a high hatefulness prediction from comment-only Bert-HateXplain and a moderate prediction from the Graph methods. However, the hate predictions from the graph methods quickly neutralize as the conversation leads into a discussion of his latest song, Montero (depths 5 and 6). Given this context, it is clear to these methods that the conversation is not hateful but rather just discussing the lyrics of the song. However, this context is not available to comment-only methods. The inaccuracy introduced without this important context can be seen by the comment-only methods, in which both systems predicted high hateful scores for both comments. This illustrates the ability of graph methods to maintain conversational context when predicting hate scores. 

\begin{table*}
    \centering
        \caption{Conversation on /r/MensRights containing contextual affiliation-directed hate speech}
    \begin{tabular}{c|p{8cm}|c|c|c|c}
        Depth & Text & Graphormer & GAT & Bert-HateXplain & MuRIL \\
        \hline
        \hline
        0 & How to get banned from r/Feminism [image] & 3 & 3 & 0 & 0 \\
        \hline
        1a & i guess they were on their period and they want their feelings to matter at that point and you got banned for it & 3 & 3 & 1 & 0  \\
        \hline 
        1b & Feminism is cancerous anyways & 3 & 2 & 1 & 0  \\
        \hline
        1c & Wow, they are almost as fragile as the donald. I guess all special snow flakes need their own safe space. & 1 & 2 & 1 & 0 \\
        \hline
        1d & Absolutely horrible. & 1 & 0 & 1 & 1 \\
    \end{tabular}
    \label{tab:contextual-mens-rights}
\end{table*}
Next, we turn to an example of a discussion where context is needed to detect hate speech. \edit{For this, we focus on the subreddit /r/MensRights.  This community advocates for increased men's rights by discussing social issues that adversely impact them, which frequently devolves into harmful misogyny \cite{socsci5020018}}.

This pattern of abuse can be seen in the discussion presented in Table \ref{tab:contextual-mens-rights}. Here, the user posts an image of a brief exchange they had in the r/Feminism subreddit. In this exchange, the user was banned from that community for advocating against the right for women to \textit{feel} safe but rather that women only have the right to \textit{be} safe, regardless of reassurances and comfort. By posting this exchange on r/MensRights, the user aimed to frame the r/Feminism community in a negative light by stating that they were banned in response to sharing a valid point. 

In this discussion, graph methods were able to accurately understand the negative context of feminism towards the hateful comments that followed. Most surprisingly, none of the comments that followed were labelled as hate speech by the comment-only methods. Such examples include "Feminism is cancerous anyways", which was appropriately labelled as hateful by graph methods, receiving a prediction of 3 from Graphormer, but mislabeled as innocuous by text methods. We assume that this large difference in prediction comes uniquely from the graph structure of the discussion. 

\begin{table*}
    \centering
        \caption{Conversation on r/rupaulsdragrace containing reclaimed language and multi-modal context}
    \begin{tabular}{c|p{8cm}|c|c|c|c}
        Depth & Text & Graphormer & GAT & Bert-HateXplain & MuRIL \\
        \hline
        \hline
        0 & *SPOILERS* Always and Forever, paparazzi who? [image - Figure \ref{fig:drag}] & 1 & 1 & 1 & 0 \\
        \hline
        1a & Am I the only f*ggot that LIVED for this look? & 1 & 2 & 4 & 3  \\
        \hline 
        2a & I honestly truly thought it was an amazing concept and I love the final result & 0 & 0 & 1 & 0 \\
        \hline
        2b & Not at all. F*ck fashion. I want fashion I'll by the Vogue fall guide. Give me something creative I haven't seen before. & 0 & 0 & 2 & 4\\
        \hline
        1b & Say what you will about the look, but can we appreciate the fact that this b*tch got over a dozen Canon DSLRs plus the lenses on this dress? That shit ain't cheap. & 1 & 1 & 3 & 4 \\
        \hline
        1c & This look is ridiculous. I love it, but my God, who thinks of this shit. & 0 & 0 & 1 & 4
    \end{tabular}
    \label{tab:contextual-drag}
\end{table*}

\begin{figure}
    \centering
    \includegraphics[width=0.55\linewidth]{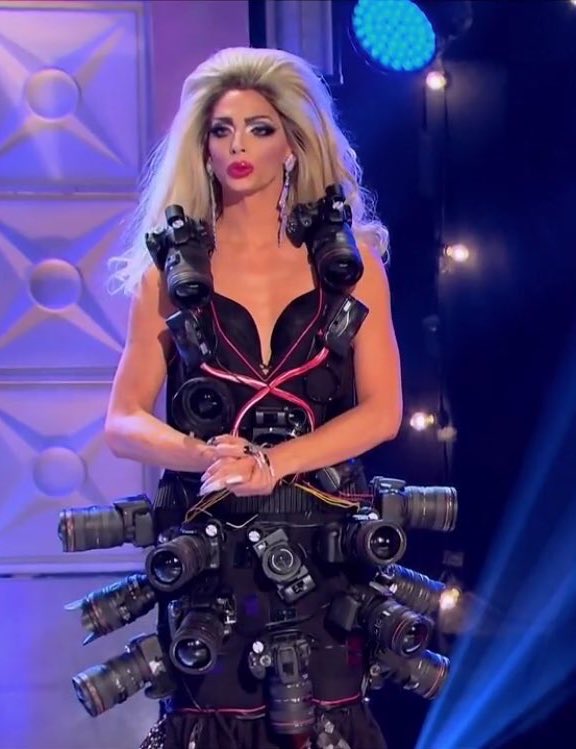}
    \caption{Photo of the Drag Queen discussed in Table \ref{tab:contextual-drag}}
    \label{fig:drag}
\end{figure}

Finally, we examine a conversation that contains reclaimed language that was previously perceived as harmful to a given community. For this, we sample a discussion from /r/rupaulsdragrace, a community dedicated to the LGBTQ drag competition Ru Paul's Drag Race (Table \ref{tab:contextual-drag}). In this conversation, users are commenting on one competitor's manufactured drag outfits. In this community, slurs such as "f*ggot" and "b*tch" are reclaimed as positive terms to refer to LGBTQ members and competitors\footnote{A common slogan on the show is "Yass b*tch", which is used to cheer on competitors}. However, these slurs are more often used in a hateful context, proving a challenge for methods that do not consider specific contexts. 

Analyzing the predictions of each method, we see that comment-only methods assign a high hate score for comments at depths 1a and 2b. However, both comments are in fact positive and supportive of the competitor mentioned in the initial post. Inspecting the content of these comments, we can infer that the false positive prediction can likely be attributed to the usage of reclaimed slurs without contextualizing them to the community and prior discussion. This behaviour is contrasted with graph methods, which predicted more accurate scores, likely due to the context provided by other comments concerning the fashion of the dress.

It is important to note that both graph and comment-only methods can only infer context by inspecting the text of comments. In this conversation, graph methods were likely only able to infer context due to the other comments, which discuss the fashion of the dress (depths 1b and 1c).  However, upon looking at the initial post, we see that the user accompanied their post with an image of a drag queen (Figure \ref{fig:drag}). This picture provides immense context to the discussion, such as the focus on fashion and the LGBTQ community. As such, we hypothesize that future work that would include multi-modal posts could provide important context to the comments that follow. 

\subsection{Inciteful Hate Speech}
\begin{table*}
    \centering
        \caption{Conversation from r/conservatives containing inciteful speech regarding Black Lives Matter \\}
    \begin{tabular}{c|p{8cm}|c|c|c|c}
        Depth & Text & Graphormer & GAT & Bert-HateXplain & MuRIL \\
        \hline
        \hline
        0 & Black Guns Matter, as does election integrity [image] & 3 & 1 & 0 & 1 \\
        \hline
        1 & That's Hilarious. I think its interesting how the left keeps trying to rub "BLACK PEOPLE ARE BUYING GUNS" in the faces of conservatives, like we would somehow be opposed to that. [...] glad that gun ownership is expanding among all demographics & 4 & 3 & 1 & 2  \\
        \hline 
        2a & It's on every conservative news and media platform as a big positive, but the left doesn't pay attention to that literally all they think is conservatives racist, therefore black guns bad for them. & 3 & 2 & 1 & 2 \\
        \hline
        2b & They want us to be divided by race like they are. They are so racist they can't imagine us being united by our love of fundamental rights & 4 & 3 & 0 & 0\\
        \hline
          & [...] & & \\
        \hline
        4 & I agree. The black panthers had the right idea. BLM members should arm too. & 2 & 0 & 0 & 0 \\
        \hline
        5 & They have been.... smh & 4 & 1 & 0 & 0\\
    \end{tabular}

    \label{tab:predictive-conservative}
\end{table*}
For our next set of examples, we investigate the ability of graph networks to predict the direction of conversations. We start by analyzing a conversation that took place on /r/conservatives, a community for discussing right-wing policies (Table \ref{tab:predictive-conservative}). These policies often include advocacy for gun rights and strong support for election denial \cite{block_2021}. Discussions in these political communities have become increasingly polarized and hateful towards members of the opposite party \cite{Waller2021}.

In this discussion, users are referring to a tweet from a black user concerning the difficulty for people to purchase weapons due to a lack of governmental IDs. The conversation is centred around confounding the Black Lives Matter movement with gun advocacy and election denial ("Black Guns Matter, as does election integrity"). The discussion then devolves into affiliation-based hate as users claim left-leaning users are racists and associate black activist groups ("the black panthers" and "BLM") with armed violence. 

Looking at the predictions of both groups of methods, we see that both graph methods predict very high hatefulness scores for many of the comments within the discussion. This can especially be seen in the comments at depth 2, in which users delve into race accusations. However, it is important to note that these comments are more akin to debate rather than explicit hate speech. This is reflected in the predictions from the comment-only methods, which consistently predict low hatefulness scores for these comments. As such, it can be inferred that these high predictions originate from a belief that the conversation will head in a hateful direction given the context thus far. Indeed, this is the case as later comments (depth 5) intensify the discussion and accuse members of the Black Lives Matter movement of armed violence. However, it is also important to note that the earlier predictions appear to conflate polarizing political discourse with hate (depth 1) with minimal discussion context. 

\begin{table*}
    \centering
        \caption{Conversation from r/politics requiring long range forecasting from community cues}
    \begin{tabular}{c|p{8cm}|c|c|c|c}
        Depth & Text & Graphormer & GAT & Bert-HateXplain & MuRIL \\
        \hline
        \hline
        0 & Trump, who was impeached for withholding nearly \$400 million in military aid from Ukraine, said 'this deadly Ukraine situation would never have happened' if he were in office [article] & 2 & 2 & 0 & 0 \\
        \hline
        1 & This happened under Biden's watch. That is a fact & 3 & 3 & 0 & 0  \\
        \hline 
        2a & Russia has been threatening Ukraine for the last 8 years & 2 & 0 & 0 & 0 \\
        \hline
        3 & Whatever, still happened under Biden's watch. Not Trump's. & 3 & 2 & 0 & 0\\
        \hline
        4 & Right, I'm sure the situation would've been so much better under the leadership of a failed jackass lapdog for Putin. & 4 & 3 & 3 & 0\\
        \hline
        2b & You might want to do a little reading about Ukraine. Your comment is completely ludicrous & 3 & 3 & 0 & 0
    \end{tabular}
    \label{tab:predictive-trump}
\end{table*}
To further examine the ability of Graphormer to predict the direction of conversations, we investigate a conversation from /r/politics. This community is known to have strong left-leaning political views and to have a distinct and polarized user base from /r/conservatives \cite{Waller2021}. Inside the sampled conversation (Table \ref{tab:predictive-trump}), users are discussing comments made by former president Trump in relation to the Ukraine war. The conversation begins benign but becomes combative,  
with one user shifting the blame for the Ukrainian war onto the current left-leaning president. This trend culminates into a climax at depth 4, where the user escalates to using hateful language against Trump. 

Investigating the predictions, we see a similar trend from the previous example where each of the predictions from the comment-only methods remains mostly neutral apart from the hateful comment at depth 4. However, the contentiousness of the conversation given the previous instigating comments is captured by graph methods, resulting in high predictions by both methods. This can especially be seen with the prediction at depth 1, which seemingly captured the contentious relationship between comments concerning Biden and Trump. Indeed, we can see that the conversation did turn hateful later in the conversation, validating this prediction.   

\section{Discussion}
In our analysis, we focused on analyzing two types of difficult hate speech: contextual and inciteful. Contextual hate speech requires conversational context to understand, and inciteful hate speech is not inherently hateful but is designed to incite further hateful comments. Both types of speech present difficulties for current comment-only approaches due to the heavy reliance on the context in order to make correct predictions. 

Starting with contextual hate speech, we analyzed three different conversations originating from /r/gay, /r/MensRights, and /r/rupaulsdragrace. Using comment-only methods, we found many predictions that were false positives or false negatives depending on the text of the comment in isolation. For false positives, we found that comment-only methods tended to predict high hate scores for comments that contained slurs (Table \ref{tab:contextual-gay} and Table \ref{tab:contextual-drag}). However, upon reading the rest of the conversation, it becomes clear that many of these slurs are utilized in a non-derogatory context. The same can be said for false negatives, where antagonistic replies can lose their hateful context when considered in isolation (Table \ref{tab:contextual-mens-rights}). However, in each of these conversations, we found that graph methods perform well at capturing the vital discussion context that is required to appropriately understand these comments. Between the two graph methods we evaluated, both GAT and Graphormer performed similarly well in the examples we explored. 

To examine the ability of graph and comment-only methods to capture inciteful speech, we analyzed two discussions from /r/conservatives and /r/politics, communities with polarizing user-bases \cite{Waller2021}. We found that graph methods are sensitive to counter-speech as evidence of inciting hateful discourse. This can be especially seen in Table \ref{tab:predictive-trump}, where users disagreeing on the cause of the Ukraine war lead to a high hatefulness prediction from the two graph methods. While there may be some validity to these predictions regarding their contentiousness, it does raise a concern about how to moderate these heated debates. 
However, in the case of the example in Table \ref{tab:predictive-conservative}, the most inciteful comments (depth 2a, 2b, and 5) are appropriately labelled as such. In each of these cases, comment-only methods predicted each comment as neutral, even if the comments were hateful (depth 2b). We found that Graphormer was more sensitive to higher predictions than GAT when faced with these types of comments.  

In each of the examples, we also evaluated the ability of graph networks to adapt to evolving social media conversations. Coping with this dynamic change is essential for the successful real-life implementation of AI for social impact. We evaluate this behaviour by iteratively predicting comments in the discussion graph in a depth-wise fashion, differing from \citet{hebert2022predicting}. As a result, we constrain graph models to predict labels from only the context provided by previous comments, mirroring how the system would be deployed in real situations. Despite this constraint, we still see that graph systems are able to make accurate predictions. This can best be seen in Table \ref{tab:contextual-gay}, where the graph models adapted their predictions to be less hateful once the conversation developed. 

We also found that many examples we retrieved were entered around multi-modal posts. Such examples include the discussion in Table \ref{tab:contextual-gay}, involving an image of a tweet, and the discussion in Table \ref{tab:predictive-trump}, involving an article concerning Trump and the Ukraine war, among others. When investigating contextual hate speech, Table \ref{tab:contextual-drag} presents an example where the image (Figure \ref{fig:drag}) provides important context to the comments that followed. 

\begin{figure}
    \centering
    \includegraphics[width=\linewidth]{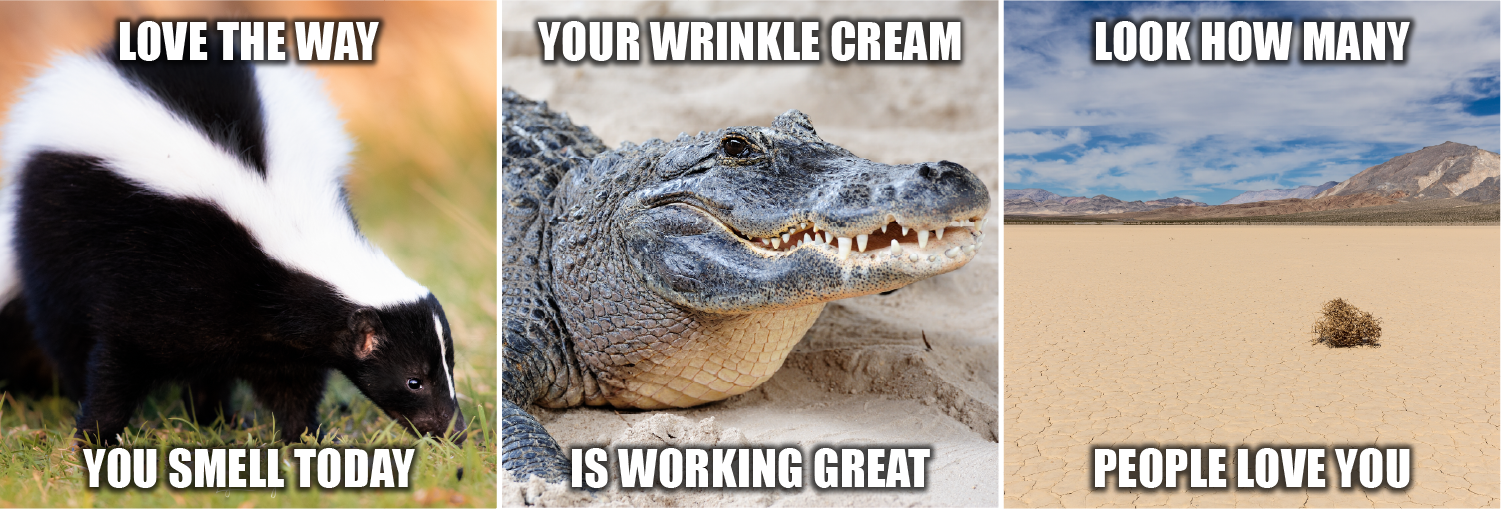}
    \caption{Examples of Multi-Modal Hate Speech}
    \label{fig:multimodal}
\end{figure}

By analyzing this picture, it would be possible to understand that the discussion concerns an LGBTQ drag queen competing in an elaborate dress. However, without this context, we found that comment-only methods misclassified supportive speech using reclaimed LGBTQ vernacular as hateful. This is especially concerning given that these predictions could serve to suppress communities that are vitally important to the mental health of minority populations \cite{lucero2017safe, fish2020stuck}. Furthermore, memes sent on online platforms are often only hateful if one considers both the image and the text caption together, as seen in Figure \ref{fig:multimodal} \cite{kiela2020hateful}. By taking a holistic view of conversations by encoding images, text, and discussion structure together, we hypothesize those hate speech detection methods would be able to avoid many false predictions, such as the ones incurred in Table \ref{tab:contextual-drag}. \edit{Furthermore, following \citet{tian-etal-2022-duck}, it would be possible to include user-level information into this graph representation.}

Finally, it is also important to analyze the mental health impact given by a graph approach to hate speech. By reformulating hate speech as a graph prediction task, we are able to train systems that can leverage discussion context toward predicting the direction of conversations. This can allow moderators on social platforms to be alerted of potentially harmful comments and deploy mitigation strategies to shield users who are susceptible to mental health effects. We see an example of such a discussion in Table \ref{tab:predictive-conservative}, where users that are susceptible to trauma from guns and race can be warned ahead of time by utilizing the proactive graph predictions. Furthermore, by utilizing an increasing ordinal scale (from zero to four) for predicting hate, users can select their level of comfort by choosing the intensity of contentious comments they are comfortable viewing. As the conversation develops, these predictions can then be updated with further context and revised accordingly. An example of where this would be useful is the discussion in Table \ref{tab:contextual-drag}, where further comments add credence to the innocence of previous comments. \edit{By providing these scores, platform owners can allow users to have control over the content they see through self-moderation. Another valued opportunity for deployment of our methods shown by qualitative analysis is in assiting platforms to curtail hate speech proliferation: greater prediction of impending escalating harm and caution in imposing penalites when discussion isn't hate can both be addressed. }      
    
\section{Conclusion}
In this work, we explored the impact of Graph Transformer Networks on hate speech detection \cite{hebert2022predicting}. To do this, we performed an extensive qualitative analysis of graph and comment-only methods on conversations sampled from different communities on Reddit. When examining contextual hate speech, we found that Graph Transformer Networks can prevent both false positives and false negatives incurred by comment-only methods. In these cases, context played a key role in understanding the nature of analyzed comments. We also found similar gains in performance when analyzing discussions that concerned inciteful speech. However, we also found that debates were prone to high hate predictions despite being mostly civil. 

Guided by this study, one promising direction for future work is to include more modalities to better contextualize comments. Among the examples we retrieved, many were centred around an image or article. We hypothesize that utilizing a holistic view of conversations by including all modalities can help prevent false positives. Most importantly, this approach could help catch the most pervasive hate speech of all - discourse.

\bibliography{aaai22}
\end{document}